\documentclass[10pt]{article}
\usepackage{setspace}
\usepackage[margin=1.1in]{geometry}

\usepackage[font=footnotesize]{caption}

\usepackage{amsgen}
\usepackage{amsfonts}
\usepackage{amssymb}
\usepackage{amsbsy}
\usepackage{graphicx}
\usepackage[breaklinks=true,colorlinks=true,linkcolor=blue,urlcolor=blue,citecolor=blue]{hyperref}
\usepackage[ruled,vlined]{algorithm2e}
\usepackage{algorithmic}
\usepackage{color}
\usepackage{multirow}

\usepackage{amsmath,amssymb}

\usepackage{caption}
\usepackage{subcaption}
\newcommand{\norm}[1]{\left\lVert#1\right\rVert}

\title{Deep Learning 3D Dose Prediction for Conventional Lung IMRT Using Consistent/Unbiased Automated Plans}
\author{Navdeep~Dahiya$^{1*}$,~Gourav~Jhanwar$^{2}$,~Anthony~Yezzi$^{1}$,~Masoud~Zarepisheh$^{2\dagger}$,~Saad~Nadeem$^{2\dagger}$}
\date{}

\usepackage{fancyhdr}
\begin{document}
\maketitle

{\footnotesize
\noindent
$^1$Department of Electrical \& Computer Engineering, Georgia Institute of Technology, Atlanta, GA, USA.

\noindent
$^2$Department of Medical Physics, Memorial Sloan-Kettering Cancer Center, New York, NY, USA.

\noindent
$^*$ Work done as an intern at MSKCC.

\noindent
$^\dagger$ Co-senior authors

\noindent
\textbf{Corresponding author:} Saad Nadeem (nadeems@mskcc.org)
}

\begin{abstract}
\noindent
Deep learning (DL) 3D dose prediction has recently gained a lot of attention. However, the variability of plan quality in the training dataset, generated manually by planners with wide range of expertise, can dramatically effect the quality of the final predictions. Moreover, any changes in the clinical criteria requires a new set of manually generated plans by planners to build a new prediction model. In this work, we instead use consistent plans generated by our in-house automated planning system (named ``ECHO'') to train the DL model. ECHO (expedited constrained hierarchical optimization) generates consistent/unbiased plans by solving large-scale constrained optimization problems sequentially. If the clinical criteria changes, a new training data set can be easily generated offline using ECHO, with no or limited human intervention, making the DL-based prediction model easily adaptable to the changes in the clinical practice. We used 120 conventional lung patients (100 for training, 20 for testing) with different beam configurations and trained our DL-model using manually-generated as well as automated ECHO plans. We evaluated different inputs: (1) CT+(PTV/OAR)contours, and (2) CT+contours+beam configurations, and different loss functions: (1) MAE (mean absolute error), and (2) MAE+DVH (dose volume histograms). The quality of the predictions was compared using different DVH metrics as well as dose-score and DVH-score, recently introduced by the {\it AAPM knowledge-based planning grand challenge}. The best results were obtained using automated ECHO plans and CT+contours+beam as training inputs and MAE+DVH as loss function.
\end{abstract}
\vspace{2pc}

%\paragraph{Significance and Novelty:} {\footnotesize ...}

\newpage

\section{Introduction}
Despite recent advances in optimization and treatment planning, intensity modulated radiation therapy (IMRT) treatment planning remains a time-consuming and resource-demanding task with the plan quality heavily dependent on the planner's experience and expertise. This problem is even more pronounced for challenging clinical cases such as conventional lung with complex geometries and intense conflict between the objectives of irradiating planning target volume (PTV) and sparing organ at risk structures (OARs).

In the last decade, Knowledge-based planning (KBP) methods have been developed to help automate the process of treatment plan generation. KBP methods represent a data-driven~\cite{KBP-Survey} approach to treatment planning whereby a database of preexisting clinical plans is utilized by predictive models to generate new patient specific plan. It involves the use of machine learning methods such as linear regression, principal component analysis, random forests, and neural networks to generate an initial plan which is then further optimized including manual input from the dosimetrist and physicians. Dose volume histogram (DVH) is a main metric used to characterize the dose distribution for given anatomical structures. The earlier KBP methods were dedicated to predicting the DVH using different underlying models/methods~\cite{DVH1,DVH2,DVH3,DVH4,DVH5,DVH6,DVH7}. DVH consists of zero-dimensional (such as mean/minimum/maximum dose) or one-dimensional metrics (volume-at-dose or dose-at-volume histograms) which lacks any spatial information. Methods based on learning to predict DVH statistics fail to take into account detailed voxel-level dose distribution in 2D or 3D. This shortcoming has led to a push towards development of methods for directly predicting voxel-level three-dimensional dose distributions.

A major driver in the push for predicting 3D voxel-level dose plans has been the advent of deep learning (DL) based methods~\cite{DL-Review}. Originally developed for tasks such as natural image segmentation, object detection, image recognition, and speech recognition, deep learning methods have found applications in medical imaging including radiation therapy~\cite{DL-Cheon2020,DL3,DL4-Jarrett2019,DL5}. A typical DL dose prediction method uses a convolutional neural network (CNN) model which receives a 2D or 3D input in the form of planning CT with OAR/PTV masks and produces a voxel-level dose distribution as its output. The predicted dose is compared to the real dose using some form of loss function such as mean squared error and gradients are backpropagated through the CNN model to iteratively improve the predictions. In recent years, many such methods have been developed using different input configurations, with different network architectures, and loss functions and have been applied to various anatomical sites including head and neck~\cite{HeadNeck1-KEARNEY2018111,HeadNeck2}, prostate~\cite{Prostate1-KANDALAN2020228,Prostate2-8983412,Prostate3-Nguyen2019,Prostate4}, pancreas~\cite{Pancreas-Wang2020FluenceMP}, breast cancer~\cite{BreastCancer1-BAKX202165}, esophagus~\cite{Esophagus1} and lung cancer sites~\cite{LungCancer1,LungCancer2,LungCancer3}. A recent review article~\cite{TreatmentPlanningReview-Wang2020} covers DL developments specifically for external beam radiotherapy automated treatment planning.

All the previous DL-based methods, however, still rely on manually-generated plans for training. A recent work~\cite{DataQuality} demonstrated the importance of consistent training data on the performance of the DL model for esophageal cancer. This work compared the performance of the same model trained on variable as well as more homogeneous/consistent plan databases. The original database contained different machines, beam configurations, beam energies and involved different physicians and medical physicists for contouring and planning respectively, whereby the homogenized/consistent version was created  by re-contouring, re-planning, and re-optimization of the plans done by the same observer with identical beam configurations. It was shown that a homogenized/consistent database led to higher performance compared to the original variable plan database. In this work, we employ our in-house automated treatment planning system, internally referred to expedited constrained hierarchical optimization (ECHO), to generate consistent high-quality plans as an input for our DL model. ECHO generates consistent high-quality plans by solving a sequence of constrained large-scale optimization problem~\cite{ECHO1-https://doi.org/10.1002/mp.13572,ECHO2-HONG20201042,ECHO3-https://doi.org/10.1002/mp.13908,ECHO4-https://doi.org/10.1002/mp.14148,ECHO5-https://doi.org/10.1002/mp.14215,ECHO6-Dursun_2021}. ECHO is integrated with Eclipse and is used in our daily clinical routine, with more than 4000 patients treated to date. The integrated ECHO-DL system proposed in this work can be quickly adapted to the clinical changes using the complementary strengths of both our ECHO and DL modules, i.e. consistent/unbiased plans generated by ECHO and the fast 3D dose prediction by the DL module.

\section{Materials and Method}
\subsection{Patient Dataset}
We use a database of 120 randomly selected lung cancer patients treated with conventional IMRT with 60 Gy in 30 fractions at Memorial Sloan Kettering Cancer Center between the year 2018 and 2020. All these patients received treatment before clinical deployment of ECHO for lung disease site and therefore include the treated plans which were manually generated by planners using 5--7 coplanar beams and 6 MV energy. We ran ECHO for these patients using the same beam configuration and energy. ECHO solves two constrained optimization problems where the critical clinical criteria in Table \ref{table:1} are strictly enforced by using constraints, and PTV coverage and OAR sparing are optimized sequentially. ECHO can be run from Eclipse\textsuperscript{TM} as a plug-in and it typically takes 1--2 hour for ECHO to automatically generate a plan. ECHO extracts the data needed for optimization (e.g., influence matrix, contours) using Eclipse\textsuperscript{TM} application programming interface (API) and solves the resultant large-scale constrained optimization problems using commercial optimization engines (KNITRO\textsuperscript{TM}/AMPL\textsuperscript{TM}) and then imports the optimal fluence map into Eclipse for final dose calculation and leaf sequencing.

\begin{table}[h!]
\caption{Clinical Max/Mean dose (in Gy) and Dose-volume criteria}
\label{table:1}
\centering
 \begin{tabular}
 {|l |c | c| c|} 
 \hline
 \textbf{Structure} & \textbf{Max (Gy)} & \textbf{Mean (Gy)} & \textbf{Dose-volume} \\ [0.5ex] 
 \hline
 PTV & 72 &  &  \\ 
 Lungs Not GTV	& 66 & 21 & V(20Gy) $<=$ 37\%\ \\
 Heart & 66 & 20 & V(30Gy) $<=$ 50\%\ \\
 Stomach & 54 & 30 & \\
 Esophagus & 66 & 34 & \\
 Liver & 66 &  & V(30Gy) $<=$ 50\%\ \\ 
 Cord & 50 & & \\
Brachial Plexus	& 65 &  & \\ 
 \hline
\end{tabular}
\end{table}

\subsection{Inputs and Preprocessing}
\begin{figure}[htb!]
    \centering
    \includegraphics[width=\linewidth]{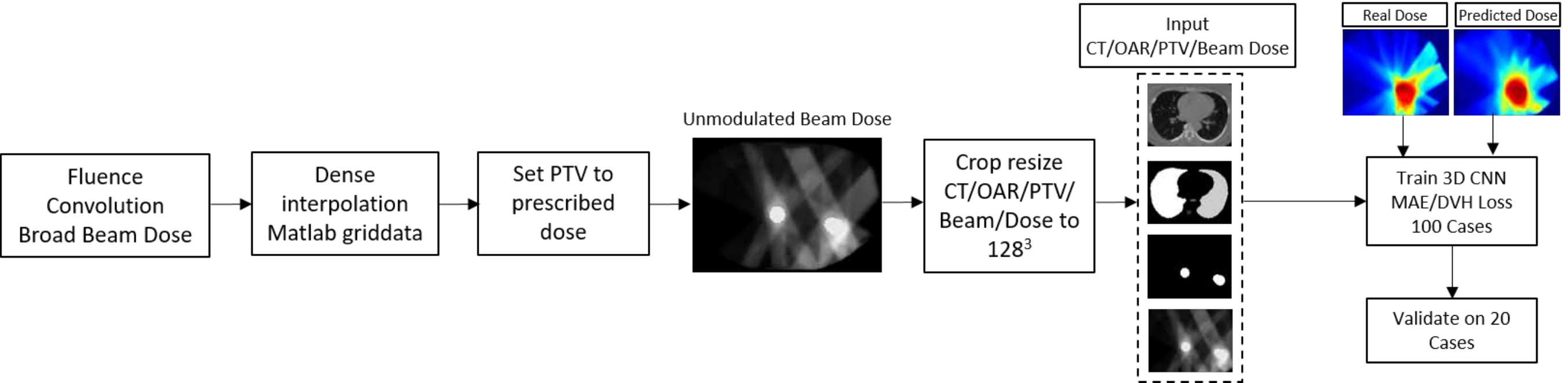}%
    \caption{Entire process of training a 3D CNN network to generate a 3D voxelwise dose. OARs are one-hot encoded and concatenated along the channel axis with CT, PTV and FCBB beam dose as input to the network.}
    \label{fig:Pipeline}
\end{figure}

Structure contours and the 3D dose distribution corresponding to the treated manual plans and automated ECHO plans were extracted from the Eclipse V15.5 (Varian Medical Systems, Palo Alto, CA, USA). Each patient has a planning CT, and corresponding PTV and OARs manual delineations which may differ from patient to patient depending on the location and size of the tumor. However, all patients have esophagus, spinal cord, heart, left lung, right lung and PTV delineated. Hence, we use these five OARs and PTV as inputs in addition to the planning CT. Similar to recent works~\cite{LungCancer1,Jiang1mp.13953}, we also incorporate the beam configuration information in the input using the fluence-convolution broad beam (FCBB) algorithm~\cite{FSBB_Lu_2010} to generate an approximate dose distribution quickly. Figure~\ref{fig:Pipeline} shows the overall workflow to train a CNN to generate voxel-wise dose distribution.

The CT images may have different spatial resolutions but have the same in-plane matrix dimensions of 512$\times$512. The PTV and OAR segmentation dimensions match those of the corresponding planning CTs. The intensity values of the input CT images are first clipped to have range of [-1000, 3071] and then rescaled to range [0, 1] for input to the DL network. The OAR segmentations are converted to a one-hot encoding scheme with value of 1 inside each anatomy and 0 outside. The PTV segmentation is then added as an extra channel to the one-hot encoded OAR segmentation.

The manual and ECHO dose data have different resolutions than the corresponding CT images. Each pair of the manual and ECHO doses is first resampled to match the corresponding CT image. The dose values are then clipped to values between [0, 70] Gy. For easier training and comparison between different patients, the mean dose inside PTV of all patients is rescaled to 60 Gy. This serves as a normalization for comparison between patients and can be easily shifted to a different prescription dose by a simple rescaling inside the PTV region. We set all the dose values inside the PTV to the prescribed dose of 60 Gy and then resample these to match the corresponding CT, similar to the original manual/ECHO doses.

Finally, in order to account for the GPU RAM budget, we crop a 300$\times$300$\times$128 region from all the input matrices (CT/OAR/PTV/Dose/Beam configuration) and resample it to a consistent 128$\times$128$\times$128 dimensions. We used the OAR/PTV segmentation masks to guide the cropping to avoid removing any critical regions of interest.

\subsection{CNN Architecture}
We train a Unet like CNN architecture~\cite{ronneberger2015unet,pix2pix2017} to output the voxel-wise 3D dose prediction corresponding to an input comprising of 3D CT/contours  and beam configuration all concatenated along the channel dimension. The network follows a common encoder-decoder style architecture which is composed of a series of layers which progressively downsample the input (encoder), until a bottleneck layer, where the process is reversed (decoder). Additionally, Unet-like skip connections are added between corresponding layers of encoder and decoder. This is done to share low-level information between the encoder and decoder counterparts.

\begin{figure}[htb!]
    \centering
    \includegraphics[width=\linewidth]{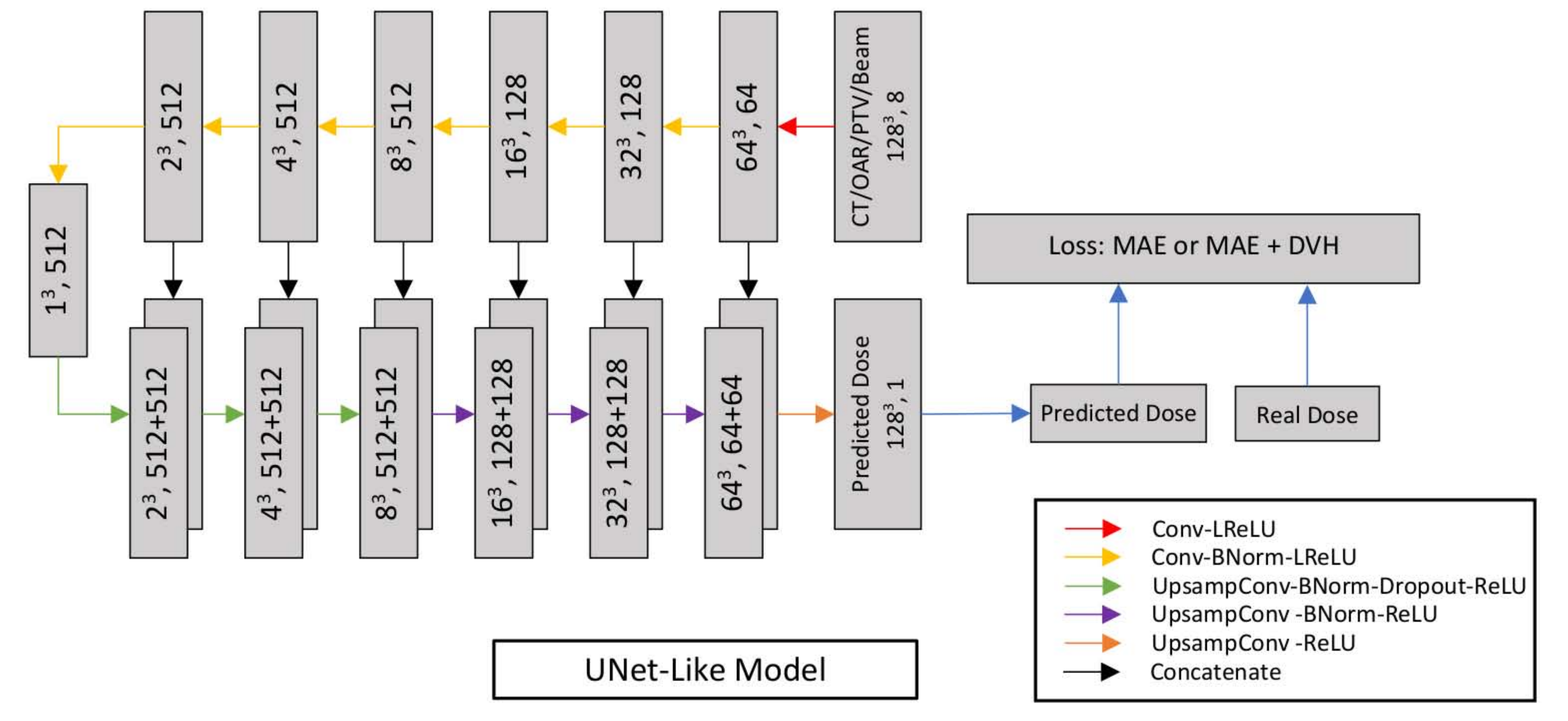}%
    \caption{A 3D Unet-like CNN architecture used to predict 3D voxelwise dose.}
    \label{fig:Unet-Like-Network}
\end{figure}

The network (Figure~\ref{fig:Unet-Like-Network}) uses combinations of Convolution-BatchNorm-ReLU and Convolution-BatchNorm-Dropout-ReLU layers with some exceptions. Batchnorm is not used in the first layer of encoder and all ReLU units in the encoder are leaky with slope of $0.2$ while the decoder uses regular ReLU units. Whenever dropout is present, a dropout rate of $50\%$ is used. All the convolutions in the encoder are $4\times4\times4$ 3D spatial filters with a stride of 2 in all 3 directions. The convolutions downsample by $2$ in the encoder. In the decoder we use trilinear upsampling followed by regular $3\times3\times3$ stride 1 convolution. The last layer in the decoder maps its input to a one channel output ($128^3, 1$) followed by a ReLU non-linearity which gives the final predicted dose. \textbf{The complete code for our DL implementation will be released via GitHub.}

\subsection{Loss Functions}
We use two types of loss functions in our study. First, we use mean absolute error (MAE) as loss function which measures the error between paired observations, which in our case are the real and predicted 3D dose. MAE is defined as $\frac{1}{N}\sum_i|D_p(i) - D_r(i)|$ where $N$ is the total number of voxels and $D_p$, $D_r$ are the predicted and real doses. We preferred to use MAE versus a common alternative, mean squared error (MSE), as MAE produces less blurring in the output compared to MSE~\cite{pix2pix2017}.

Recent work~\cite{DVHLoss} has shown the importance of adding a domain-knowledge loss function based on DVH along with MAE. The non-differentiability issue~\cite{zhang_2018} of DVH can be addressed by approximating the heaviside step function by the readily differentiable sigmoid function. The volume-at-dose with respect to the dose $d_t$ is defined as the volume fraction of a given region-of-interest (OARs or PTV) which receives a dose of at least $d_t$ or higher. Borrowing the notation from~\cite{DVHLoss}, given a segmentation mask, $M_s$, for the $sth$ structure, and a volumetric dose distribution, $D$, the volume at or above a given threshold, $d_t$, can be approximated as:
\begin{equation}
    v_{s,t}(D,M_s) = \frac{\sum_i \sigma\left( \frac{D(i) - d_t}{\beta} \right)M_s(i)}{\sum_iM_s(i)}
\end{equation}
where $\sigma$ is the sigmoid function, $\sigma(x) = \frac{1}{1 + e^{-x}}$, $\beta$ is histogram bin width, and $i$ loops over the voxel indices of the dose distribution. The DVH loss can be calculated using MSE between the real and predicted dose DVH and is defined as follows:
\begin{equation}
    L_{DVH}\left( D_r, D_p, M\right) = \frac{1}{n_s} \frac{1}{n_t} \sum_s \norm{DVH\left(D_r,M_s\right) - DVH\left(D_p, M_s\right)}_2^2
\end{equation}

\subsection{Evaluation Criteria}
To evaluate the quality of the predicted doses, we adopt the metrics used in a recent AAPM ``open-access knowledge-based planning grand challenge'' (OpenKBP). This competition was designed to advance fair and consistent comparisons of dose prediction methods for knowledge-based planning in radiation therapy research. The competition organizers used two separate scores to evaluate dose prediction models: dose score, which evaluates the overall 3D dose distribution and a DVH score, which evaluates a set of DVH metrics. The dose score was simply the MAE between real dose and predicted dose. The DVH score which was chosen as a radiation therapy specific clinical measure of prediction quality involved a set of DVH criteria for each OAR and target PTV. Mean dose received by OAR was used as the DVH criteria for OAR while PTV had three criteria: D1, D95, and D99 which are the doses received by 1\% ($99^{th}$ percentile), 95\% ($5^{th}$ percentile), and 99\% ($1^{st}$ percentile) of voxels in the target PTV. DVH error, the absolute difference between the DVH criteria for real and predicted dose, was used to evaluate the DVHs. Average of all DVH errors was taken to encapsulate the different DVH criteria into a single score measuring the DVH quality of the predicted dose distributions.

We also report additional DVH metrics for different anatomies typically used in clinical practice to evaluate dose plans. D2, D95, D98, D99 are radiation doses delivered to 2\%, 95\%, 98\% and 99\% of the volume and calculated as a percentage of the prescribed dose (60 Gy). Dmean (Gy) is the mean dose of the corresponding OAR again expressed as a percentage of prescribed dose. V5, V20, V35, V40, V50 are the percentage of corresponding OAR volume receiving over 5 Gy, 20 Gy, 35 Gy, 40 Gy and 50 Gy respectively. We report the MAE (mean $\pm$ STD) between the ground truth and predicted values of these metrics.

\subsection{Deep Learning Settings}
In our experiments, used to report the final results, we use Stochastic Gradient Descent (SGD), with a batch size of 1, and Adam optimizer~\cite{DBLP:journals/corr/KingmaB14} with an initial learning rate of $0.0002$, and momentum parameters $\beta_1 = 0.5$, $\beta_2 = 0.999$. We train the network for total of 200 epochs. We use a constant learning rate of $0.0002$ for the first 100 epochs and then let the learning rate linearly decay to 0 for the final 100 epochs. When using the MAE and DVH combined loss, we scale the DVH component of the loss by a factor of 10.

We divided our training set of 100 images into train/validation set of 80 and 20 images respectively and determined the best learning rate and scaling factor for MAE+DVH loss. Afterwards, we train all our models using all 100 training datasets and test on the holdout 20 datasets used for reporting results.

We created the implementations of the CNN model, loss functions and other related training/testing scripts in pytorch and we conducted all our experiments on a Nvidia RTX 2080 Ti GPU with 11 GB VRAM.

\section{Results}
\begin{figure}[htb!]
    \centering
    \includegraphics[width=0.95\linewidth]{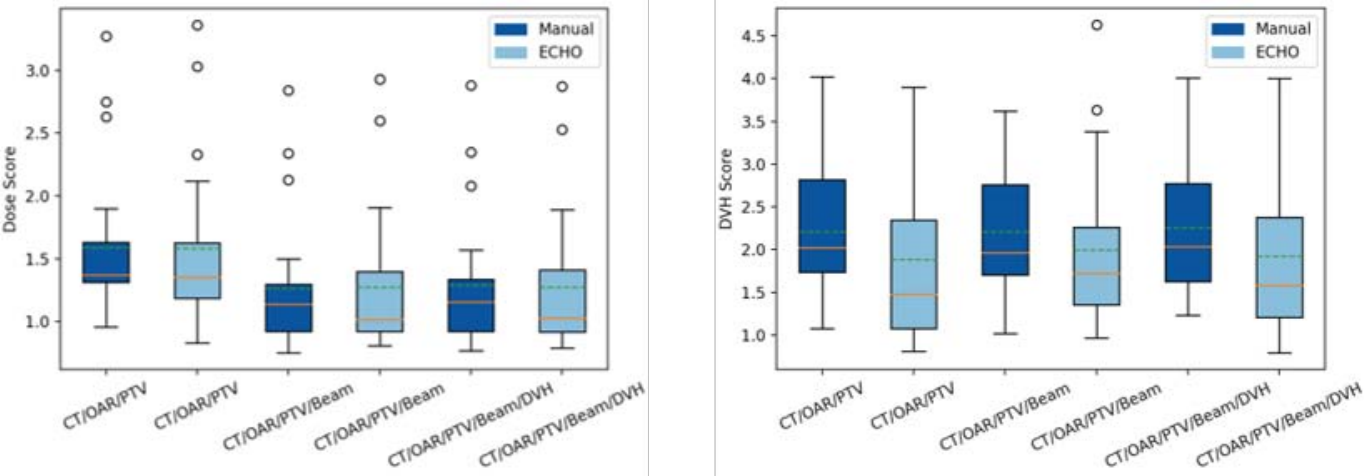}%
    \caption{Boxplots illustrating the statistics of OpenKBP dose and DVH scores for all 20 test datasets using different inputs, loss functions, and training datasets.}
    \label{fig:MetricsBoxPlots}
\end{figure}

Table~\ref{tab:OpenKBPcriteria} presents the OpenKBP metrics for 3D dose prediction using ECHO and manual training data sets with different inputs: (1) CT+Contours, and (2) CT+Contours+Beam, and different loss functions: (1) MAE, and (2) MAE+DVH. The box plot of the metrics are also provided in Figure~\ref{fig:MetricsBoxPlots} for better visual comparisons. DVH scores consistently show that the predictions for ECHO plans outperform the predictions for manual plans, whereby the dose score show comparable results. Adding beam configuration seems to improve the dose-score for both ECHO and manual plans, while adding DVH loss function only benefits the DVH-score for ECHO plans. 

\begin{table}[ht!]
\scriptsize
\begin{center}
\caption{OpenKBP evaluation metrics for various experimental settings including different inputs and loss functions to compare using ECHO vs manual plans for dose prediction.
\label{tab:OpenKBPcriteria}
\vspace*{2ex}
}
\setlength{\tabcolsep}{10pt}
\begin{tabular}{|l|c|c|c|c|}
\hline
\multirow{2}{*}{\textbf{Experiment}} & \multicolumn{2}{c|}{\textbf{ECHO}} & \multicolumn{2}{c|}{\textbf{Manual}} \\ \cline{2-5} 
                            & Dose Score (Gy)   & DVH Score (Gy)  & Dose Score (Gy)    & DVH Score (Gy)   \\ \hline
CT+Contours+Beam/MAE+DVH               & 1.27 $\pm$ 0.57         & 1.92 $\pm$ 0.92       & 1.29 $\pm$ 0.54          & 2.25 $\pm$ 0.73        \\ \hline
%MAE-DVH-Sparse              & 1.28         & 1.84       & 1.31          & 2.27        \\ \hline
CT+Contours+Beam/MAE                   & 1.28 $\pm$ 0.59         & 1.99 $\pm$ 0.91       & 1.27 $\pm$ 0.54          & 2.21 $\pm$ 0.68        \\ \hline
%MAE-Sparse                  & 1.27         & 1.92       & 1.31          & 2.27        \\ \hline
%MAE-No-Beam                 & 1.56         & 2.08       & 1.59          & 2.46        \\ \hline
CT+Contours/MAE                 & 1.58 $\pm$ 0.64         & 1.89 $\pm$ 1.02       & 1.59 $\pm$ 0.59          & 2.21 $\pm$ 0.79       \\ \hline
\end{tabular}
\end{center}
\end{table}

\begin{figure}[htb!]
    \centering
    \includegraphics[width=0.95\linewidth]{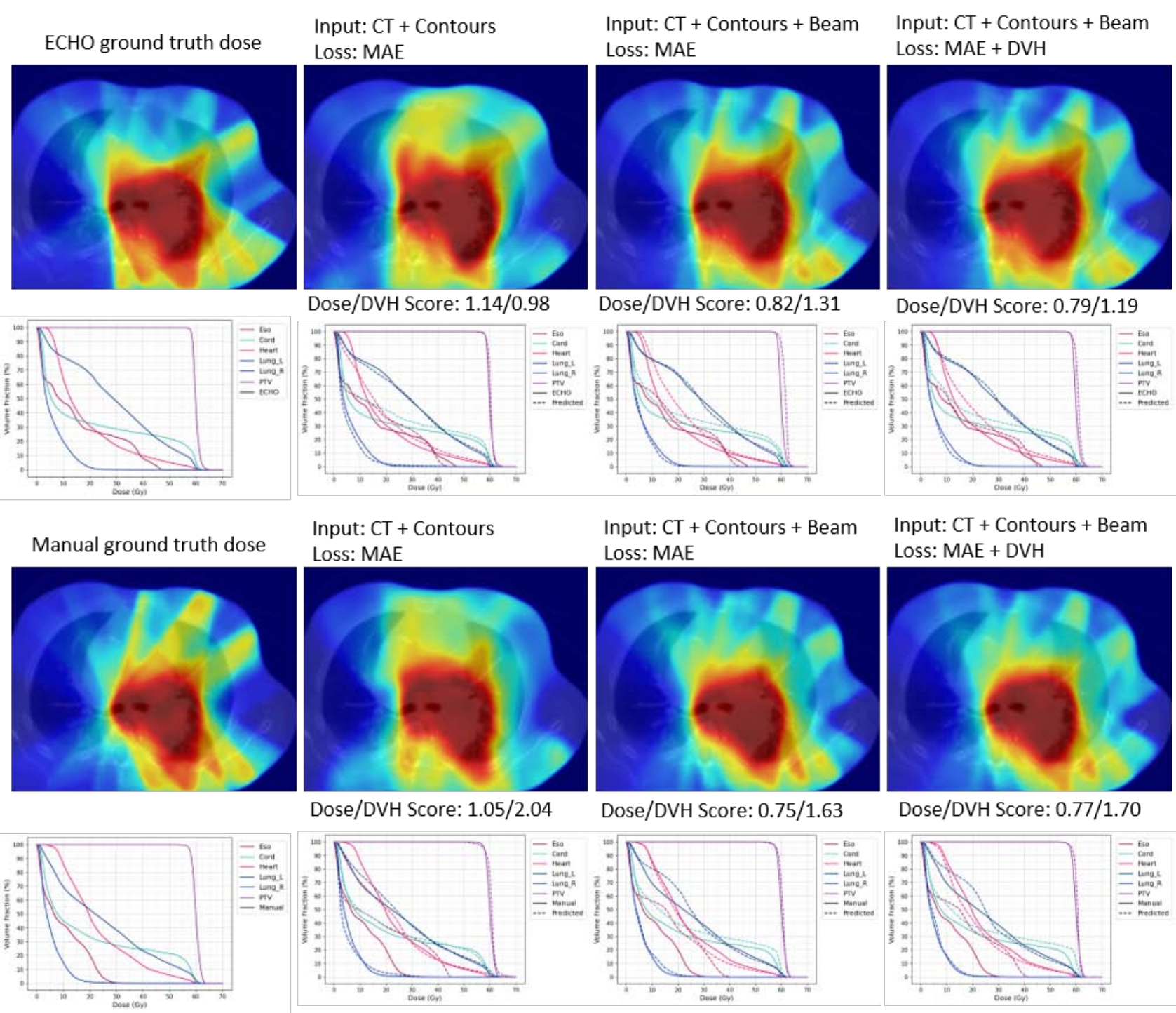}%
    \caption{An example of predicted doses using CT with OAR/PTV only, CT with OAR/PTV/Beam with MAE loss and CT with OAR/PTV/Beam with MAE+DVH loss as inputs to the CNN.}
    \label{fig:QualitativeResults1}
\end{figure}

Figure~\ref{fig:QualitativeResults1} shows an example of predicted manual and ECHO doses for the same patient using different input and loss function configurations. The dose distributions reveal the benefits of adding beams to the input. For both ECHO and manual plans, using only the CT, OAR and PTV as the input to the network produces generally blurred output dose. There is no visible dose spread in beam directions. Adding beam configuration as an extra input produces dose output which looks more similar to the real dose and spreads the dose more reliably along the beam directions. Without the beam configuration as an extra input, the DL network is unable to learn the beam structure, and simply distributes the dose in the PTV and OAR regions. It has no concept of the physics of radiation beams and when we use the beam as an extra input, we are forcing the network to learn that. DVH plots illustrate that adding DVH loss function slightly improves the ECHO prediction while it degrades performance for the manual plan prediction. Looking at both dose distribution and DVH, the best manual/ECHO results are obtained using all the inputs (CT+Contours+Beams), while adding DVH loss function only benefits ECHO. 

Table~\ref{tab:DVHMertics} compares predictions of ECHO and manual plans using different configurations and clinically-relevant metrics. Again, in general, the best result is obtained when the network is trained using ECHO plans with all the inputs and MAE+DVH as the loss function.

\begin{table}[ht!]
\tiny
\begin{center}
\caption{Mean absolute error and its standard deviation (mean $\pm$ std) for relevant DVH metrics on PTV and several organs for the test set using manual and ECHO data with (a) CT+Contours/MAE, (b) CT+Contours+Beam/MAE, and (c) CT+Contours+Beam/MAE+DVH combinations. The values are expressed as percentage of the prescription dose ($D_{pre} = 60$ Gy) for the metrics reporting the dose received by $x\%$ of volume ($D_x$), and as an absolute difference for the metrics reporting the volume (in $\%$) receiving a dose of $y$ Gy. 
\label{tab:DVHMertics}
}
\setlength{\tabcolsep}{3pt}
\begin{tabular}{|l|c|c|c||c|c|c|}
\hline
\multirow{2}{*}{} & \multicolumn{3}{c||}{\textbf{ECHO}} & \multicolumn{3}{c|}{\textbf{Manual}} \\ \cline{2-7} 
& CT+Contours & CT+Contours+Beam & CT+Contours+Beam & CT+Contours & CT+Contours+Beam & CT+Contours+Beam \\ 
& MAE & MAE & MAE+DVH & MAE & MAE & MAE+DVH \\
\hline

\textbf{PTV} & & & & & & \\
\qquad $D_{99}$ ($\%$ of $D_{pre}$)  & 4.69$\pm$6.98 & 3.85$\pm$3.08 & 3.70$\pm$3.57 & 6.24$\pm$4.95 & 6.68$\pm$4.53 & 6.37$\pm$4.93\\
\qquad $D_{98}$ ($\%$ of $D_{pre}$)  & 2.13$\pm$2.71 & 3.31$\pm$2.54 & 3.09$\pm$2.95 & 3.93$\pm$3.23 & 4.84$\pm$3.22 & 4.80$\pm$3.42 \\
\qquad $D_{95}$ ($\%$ of $D_{pre}$)  & 1.31$\pm$1.40 & 2.46$\pm$1.04 & 2.11$\pm$1.80 & 1.94$\pm$1.84 & 2.96$\pm$2.10 & 3.03$\pm$2.19\\
\qquad $D_{5}$ ($\%$ of $D_{pre}$)   & 1.15$\pm$1.96 & 1.22$\pm$1.02 & 1.51$\pm$1.02 & 1.50$\pm$1.20 & 1.52$\pm$1.28 & 1.45$\pm$1.29\\

\textbf{Esophagus} & & & & & & \\
\qquad $D_{2}$ ($\%$ of $D_{pre}$)  & 8.77$\pm$10.25 & 8.53$\pm$5.62 & 8.48$\pm$5.31 & 11.33$\pm$8.66 & 6.99$\pm$5.26 & 7.44$\pm$5.40\\
\qquad $V_{40}$ ($\%$ of volume)    & 8.78$\pm$7.94 & 9.64$\pm$9.24 & 10.58$\pm$9.16 & 12.31$\pm$8.90 & 9.34$\pm$12.09 & 10.35$\pm$11.83\\
\qquad $V_{50}$ ($\%$ of volume)    & 16.28$\pm$11.37 & 16.31$\pm$12.16 & 16.16$\pm$12.18 & 15.42$\pm$16.03 & 15.46$\pm$16.49 & 15.52$\pm$16.47\\

\textbf{Heart} & & & & & & \\
\qquad $V_{35}$ ($\%$ of volume)    & 1.57$\pm$1.43 & 2.18$\pm$1.59 & 1.90$\pm$1.71 & 2.44$\pm$2.10 & 1.99$\pm$1.75 & 1.98$\pm$1.81\\

\textbf{Spinal Cord} & & & & & & \\
\qquad $D_{2}$ ($\%$ of $D_{pre}$)  & 1.90$\pm$2.82 & 2.11$\pm$5.26 & 1.75$\pm$3.97 & 1.14$\pm$1.33 & 1.78$\pm$2.96 & 2.01$\pm$2.68\\

\textbf{Left Lung} & & & & & & \\
\qquad $D_{mean}$ ($\%$ of $D_{pre}$)  & 1.91$\pm$2.60 & 2.52$\pm$4.03 & 2.44$\pm$3.50 & 2.58$\pm$1.92 & 2.64$\pm$2.64 & 2.57$\pm$2.44\\
\qquad $V_{5}$ ($\%$ of volume)        & 4.08$\pm$3.44 & 2.81$\pm$2.30 & 2.61$\pm$2.44 & 4.23$\pm$2.84 & 3.76$\pm$3.29 & 3.94$\pm$3.70\\
\qquad $V_{20}$ ($\%$ of volume)       & 3.53$\pm$3.64 & 4.75$\pm$3.49 & 4.67$\pm$3.69 & 4.40$\pm$3.60 & 5.27$\pm$4.68 & 5.34$\pm$4.71\\

\textbf{Right Lung} & & & & & & \\
\qquad $D_{mean}$ ($\%$ of $D_{pre}$)  & 2.55$\pm$3.15 & 3.16$\pm$3.10 & 2.95$\pm$3.23 & 2.55$\pm$2.24 & 3.05$\pm$2.45 & 3.17$\pm$2.45\\
\qquad $V_{5}$ ($\%$ of volume)        & 0.01$\pm$0.03 & 0.00$\pm$0.02 & 0.01$\pm$0.03 & 0.01$\pm$0.03 & 0.00$\pm$0.02 & 0.01$\pm$0.03\\
\qquad $V_{20}$ ($\%$ of volume)       & 0.01$\pm$0.06 & 0.01$\pm$0.03 & 0.01$\pm$0.05 & 0.01$\pm$0.06 & 0.01$\pm$0.04 & 0.01$\pm$0.05\\

\hline

\end{tabular}
\end{center}
\end{table}

\section{Discussion}
This work shows an automated planning technique such as ECHO and a deep learning (DL) model for dose prediction can compliment each other. The variability in the training data set generated by different planners can deteriorate the performance of deep learning models and ECHO can address this issue by providing consistent high-quality plans. More importantly, offline-generated ECHO plans allow DL models to easily adapt themselves to changes in clinical criteria and practice. On the other hand, the fast predicted 3D dose distribution from DL models can guide ECHO to generate a deliverable Pareto optimal plan quickly; the inference time for our model is 0.4 secs per case as opposed to 1-2 hours needed to generate the plan from ECHO.

One can use an unconstrained optimization framework and penalize the deviation of the delivered dose from the predicted dose to quickly replicate the predicted plan on the new patient \cite{fan2019automatic}. However, given the prediction errors, the lack of incentive to further improve the plan if possible, and the absence of constraints to ensure the satisfaction of important clinical criteria, the optimized plan may not be Pareto or clinically optimal. A more reliable and robust approach can leverage constrained optimization framework such as ECHO. The predicted 3D dose can potentially accelerate the optimization process of solving large-scale constrained optimization problems by identifying and eliminating unnecessary/redundant constraints up front. For instance, a maximum dose constraint on a structure is typically handled by imposing the constraint on all voxels of that structure. Using 3D dose prediction, one can only impose constraints on voxels with predicted high dose and use the objective function to encourage lower dose to the remaining voxels. The predicted dose can also guide the patient's body sampling and reduce the number of voxels in optimization. For instance, one can use finer resolution in regions with predicted high dose gradient and coarser resolution otherwise.

In this work, we also investigated using different inputs and loss functions that have been previously suggested by different groups. We found that adding beams, along with CT and contours, improve the prediction for both manual and ECHO plans which is consistent with previous literature \cite{barragan2019three,zhou2020method}. For the loss function, however, our finding is only consistent with other groups \cite{DVHLoss} when DL model is trained using ECHO plans (using MAE and DVH as loss function improves the prediction). This could be due to the large variation in manual plans!

\section{Conclusion}
This work highlights the impact of large variations in the training data on the performance of DL models for predicting 3D dose distribution. It also shows that the DL models can drastically benefit from an independent automated treatment planning system such as ECHO. The consistent and unbiased training data generated by ECHO not only enhances the prediction accuracy but can also allow the DL models to be rapidly and easily adapted to the dynamic and constantly-changing clinical environments. 

\section*{Acknowledgments}
This project was supported by MSK Cancer Center Support Grant/Core Grant (P30 CA008748).

\section*{Financial Disclosures}
The authors have no conflicts to disclose.

\end{document}